\documentclass{article}
\usepackage[a4paper, total={6in,10in}]{geometry}

\usepackage[colorlinks]{hyperref}
\usepackage{amssymb,amsmath}
\usepackage[dvipsnames]{xcolor}
\usepackage{adjustbox}
\usepackage{graphicx}
\usepackage[switch,columnwise]{lineno}
\usepackage{float}

\newcommand{\norm}[1]{\left\lVert#1\right\rVert}

\newcommand{\myremark}[3]{\textcolor{#3}{\textsc{#1:}} \textcolor{#3}{\textsf{#2}}}

\newcommand{\jonathan}[1]{\myremark{Jonathan}{#1}{Red}}
\newcommand{\parker}[1]{\myremark{Parker}{#1}{Purple}}

\title{Robot Development and Path Planning for Indoor Ultraviolet Light Disinfection}

\author{Jonathan Conroy\thanks{Tufts University, MA, USA.\protect\url{{jonathan.conroy,christopher.thierauf,parker.rule,evan.krause,andrei.gonczi,matthias.scheutz}@tufts.edu}}\and
Christopher Thierauf\footnotemark[1]\and
Parker Rule\footnotemark[1]\and
Evan Krause\footnotemark[1]\and
Hugo Akitaya\thanks{University of Massachusetts Lowell, MA, USA. \protect\url{hugo_akitaya@uml.edu}}\and
Andrei Gonczi\footnotemark[1]\and
Matias Korman\thanks{Siemens Electronic Design Automation, OR, USA, \protect\url{matias_korman@mentor.com}}\and
Matthias Scheutz
}
\date{}

\begin{document}

\maketitle
\begin{abstract}
  Regular irradiation of indoor environments with ultraviolet C (UVC)
  light has become a regular task for many indoor settings as a result of
  COVID-19, but current robotic systems attempting to automate it suffer
  from high costs and inefficient irradiation. In
  this paper, we propose a purpose-made inexpensive robotic platform
  with off-the-shelf components and standard navigation software that,
  with a novel algorithm for finding optimal irradiation
  locations, addresses both shortcomings to offer affordable and
  efficient solutions for UVC irradiation.  We demonstrate in
  simulations the efficacy of the algorithm and show a prototypical
  run of the autonomous integrated robotic system in an indoor environment. In our sample instances, our proposed algorithm reduces the time needed by roughly 30\% while it increases the coverage by a factor of 35\% (when compared to the best possible placement of a static light). 
\end{abstract}

\section{Introduction}

The new coronavirus has changed our world forever. Among the many
lasting changes prompted by the rapid spread of SARS-CoV-2 is the need
for regular systematic disinfection of indoor spaces, which has gone beyond
hospitals and care facilities where room disinfection has always been
a critical task that was performed on a regular basis (e.g., when
patients were discharged) to public and private spaces
such as schools, colleges, hospitality settings,
airlines, train companies, and mass transportation authorities. In these spaces, regular disinfection will become a
critical component of any strategy to reopen societies after a
pandemic like the current one. However, this increases danger and time strain
to the human workforce that is already stretched thin. Automating these
processes is therefore an attractive approach.
In the words of UVD Robots CEO Per Juul Nielsen, whose
company builds robots with ultraviolet C (UVC) lights: ``Hospitals
around the world are waking up to autonomous disinfection. We can’t
build these robots fast
enough.''\footnote{\label{footnote:introquote}\href{https://www.forbes.com/sites/richblake1/2020/04/17/in-covid-19-fight-robots-report-for-disinfection-duty/}{www.forbes.com/sites/richblake1/2020/04/17/in-covid-19-fight-robots-report-for-disinfection-duty/}}

While solutions of this nature exist (see \ref{section:related_work}),
there are three main shortcomings of the current
solutions that severely limit their utility and applicability: (1)
cost (the existing robots are very expensive, in some cases upwards of \$100,000), (2)
formal guarantees that every surface is disinfected, and (3)
efficiency. It is unclear how well existing systems operate in part
because of the lack of formal guarantees of the disinfection process.

In this paper, we address all three aspects with the design of an
autonomous inexpensive custom robotic platform for UVC disinfection
tasks.  The robot can carry a significant payload, allowing it to
operate for extended periods of time on battery power.  More
importantly, by finding close-to-optimal paths through the indoor space, the robot minimizes power use while
guaranteeing that all surfaces receive sufficient light to guarantee disinfection.

\section{Motivation and Related Work}
\label{section:related_work}
Ultraviolet light irradiation has been used in health care settings
for quite some time to deactivate contaminants.
Initial studies suggested COVID-19 
would be similar to other corona
viruses \cite{hesslingetal20gms},
and studies have since
confirmed the exact irradiation dose necessary to deactivate the virus
(e.g., \cite{heilinglohetal20ajic}).

Existing approaches are split between
systems which are stationary in operation (e.g., Tru-D\texttrademark \ or
\cite{stationary_uvc}) and mobile platforms that have become a more
frequent research focus: consider UVD Robots\textregistered, \cite{uvc_robot1}, \cite{aircraft_sanitize}, and other ongoing projects which have seen media coverage.\footnote{See Violet (\href{https://time.com/5825860/coronavirus-robot/}{time.com/5825860/coronavirus-robot/}), work by Rovenso (\href{https://spectrum.ieee.org/automaton/robotics/industrial-robots/rovenso-uv-disinfection-robot}{spectrum.ieee.org/automaton/robotics/industrial-robots/rovenso-uv-disinfection-robot}), or the "ADAMMS-UV" from the USC Viterbi Center for Advanced Manufacturing, footnote \ref{footnote:introquote}.}
However, these platforms have either focused on the ability to bring a UVC lamp to a region rather than validating disinfection or have focused on proving disinfection experimentally
rather than formally. See \cite{trud_eval}, \cite{trud_eval2},
and \cite{ventilation_uvc_eval} as examples of stationary options, and
\cite{uvc_wand_eval} as an example of a non-stationary option using a UVC wand.

Keeping devices stationary in a
single location is suboptimal because light energy falls off with the
square of the distance (see Section~\ref{sec:model} for details) and
the farthest surface point from the device is thus the determinant of
the overall irradiation duration based on the minimum light exposure
needed to deactivate the coronavirus.

\begin{figure}[htbp]
    \centering
    \includegraphics{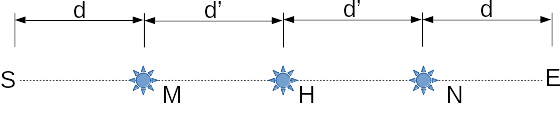}
    \caption{Example to demonstrate the need to have multiple
      locations for UVC irradiation.}
    \label{fig:example}
\end{figure}

To see this, consider a UVC lamp placement in location $H$, the
halfway point, in the simplified 1D irradiation problem in
Fig.~\ref{fig:example}. 
We assume the irradiation at all points long the line need to be at least
1. Consider sequential lamp placed in $M$ and $N$
such that the light intensity at the farthest points from the lamp $I$
at $S$, $E$, and $H$ is the same.  Since $I(S) = I(E) = 1/d^2 + 1/(d +
2\cdot d')^2$ and $I(H) = 2 \cdot 1/d'^2$, we want to find $d$ and $d'$ such
that $1/d^2 + 1/(d + 2\cdot d)^2 - 2 \cdot 1/d'^2 = 0$.  The solutions are $d =
(E-S) \cdot (3 - \sqrt 3)/6 $ and $d' = (E-S) \cdot \sqrt 3/6$ and the
overall irradiation time is $d'^2$, $d'^2/2$ each in $M$ and in $N$
(since we have to irradiate sequentially, it would be $d'^2/2$ if it
were to be done in parallel with two lamps), a significant savings
compared to $(d+d')^2$.

In real-world settings there are additional difficulties: 
(1) 2D floor plans are more complex than the 1D approximation of
a hallway provided and will therefore require more complex point
placement (see Section~\ref{section:method}), (2) robot navigation
constraints will create point placement constraints, and (3) some
obstacles may cause zones not reachable with direct light. Additionally,
the example does not address movement time as a disinfection opportunity (the lamp
can remain turned on while traveling through
its environment), and additional challenges arise from operating in
3D with a platform only capable of moving in 2D. In this paper, we
attempt to address (1), (2), and (3), while leaving the remainder for future work.

\section{Low-Cost Robot Platform}\label{section:hardware}

We developed a low-cost robotic platform equipped with an NVidia Jetson
Nano embedded computing board for onboard computing (as outlined in
Section~\ref{section:software_stack}) and a UVC germicidal
lamp (which produces 17 Watts of UV-C radiation per 100 hours) for UVC disinfection. The robot is
made of aluminium extrusion, with a square base of 24 by 28
inches. This is wide enough to provide stability for the pillar (which holds the lamp 34 inches above the ground) and space for the uninterruptible power supply (UPS) used for rechargeable power, while being thin enough to fit through a standard sized door.

\begin{figure}
    \centering
    \includegraphics[width=2in]{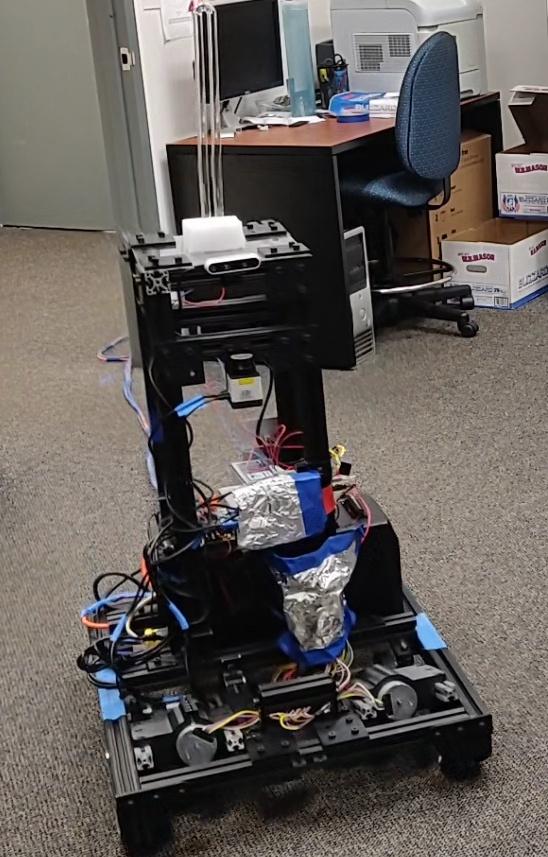}
    \caption{The robot developed for autonomous UVC irradiation
      driving into a room to be disinfected.}
    \label{fig:robot}
\end{figure}

Also mounted on the pillar are the robot's sensors. Two LIDAR units
(the Hokuyo URG 04LX) provide a 360-degree view of the environment,
which was found to be necessary to address points which were close to
obstacles by allowing the robot to safely reverse out of such
spots. The Intel Realesense D435i provides a depth point cloud of the
environment in front of the robot, allowing for obstacle avoidance of
objects that may fall above or below the linear cloud collected by the
LIDAR. Although the on-board IMU was initially used for odometery,
this was found to be unnecessary due to the accuracy of the Canonical
Scan Matcher (CSM) \cite{censi08plicp} which instead uses laser scan
matching for odometery calculation.\footnote{We have used the CCNY ROS
wrapping of CSM, available in the ROS \texttt{scan\_tools} package.}

The platform is differentially wheeled for the purposes of cost
reduction. Additional stability is provided through passive
omni-directional wheels in each corner. Both 4-inch polyurethane
wheels interface with 24-volt brushless DC motors via drivers
controlled over a USB-to-serial connection, and are on a spring-loaded
mechanism to maintain contact with uneven surfaces. Brushless DC
motors were chosen for their ability to produce high torque even at
low speeds (necessary for the stop-and-go behavior of the heavy
platform). Hall effect sensors attached to each commutator provide
wheel rotation data, which is fed into the odometry calculations.

\subsection{Navigation Software}
\label{section:software_stack}

The software running the robot is based on the Robot
Operating System (ROS \cite{ros}) and utilizes standard mapping,
control, and odometery packages in addition to some custom software
packages for hardware control. Real-time path planning is performed
using the ROS \texttt{amcl} package,
which provides an implementation of works described in
\cite{probabilisticrobotics}.
ROSControl \cite{ros_control} in
conjunction with custom hardware code manages wheel velocities when
attempting to complete a given trajectory. Mapping is performed using
GMapping, which is described in \cite{gmapping1,gmapping2}. CSM \cite{censi08plicp} is used for odometry. The
combination of these technologies enables basic obstacle-free waypoint
navigation, which is required for the proposed algorithms for finding
close-to-optimal disinfection paths described next.

\section{Planning Paths for Disinfection}
\label{section:method}

Our path planning algorithm receives the point cloud from the robot's sensors. For
tractability and modularity, we divide our algorithm into three
independent pieces: (1) {\em mapping} (given point cloud data gathered
by the robot, we obtain a floor plan of the building represented by a
simple polygon); (2) {\em waypoint and time determination} (find
locations and amount of time that the robot should stop and turn on the UVC light to so fully disinfect the floor); and (3) {\em route planning} (choose the order
in which the waypoints are visited so as to minimize travel
time).
 
Splitting the algorithm into independent subproblems may introduce
additional error. In particular, the static placement of waypoints
does not take into account the additional disinfection that may happen
while the robot moves between waypoints. This error is proportional to
the ratio between the amount of time spent moving and the time spent
at the waypoints. For our application, this ratio is quite small (we
spend hours at waypoints and a few minutes in motion). Thus, we
believe that the overall error produced is negligible.

\subsection{Mapping}

Maps are generated through usage of a custom hardware platform (see
Section~\ref{section:hardware}). Laser scan data is converted into a
2D, top-down representation of the environment in which the robot can
operate. The task of the mapping algorithm is to determine the
location of a 2D curve that can approximate this occupancy grid. The
curve represents the walls of the building (possibly with some
furniture). Knowing the exact floor plan is critical in making sure
that all locations of the room will be properly disinfected.

This problem is known as {\em curve fitting} or {\em curve
  reconstruction} in the literature (see the excellent survey by Dey
in~\cite{survey}). There are many approaches used to solve this
problem. In our setting, the curve should follow the walls and
possibly furniture of the room. A distinguishing feature from the
general problem is that walls and furniture typically are rectangular.

As the sensors used are prone to noise caused by reflections, we
preprocess the initial occupancy grid by performing a morphological
closing to remove thin areas incorrectly labeled as free space. We
then construct a polyline describing the room boundary from the
occupancy grid using the algorithm described in~\cite{contours} and
implemented in OpenCV.

\begin{figure}[htbp]
    \centering
    \includegraphics[width=0.4\columnwidth]{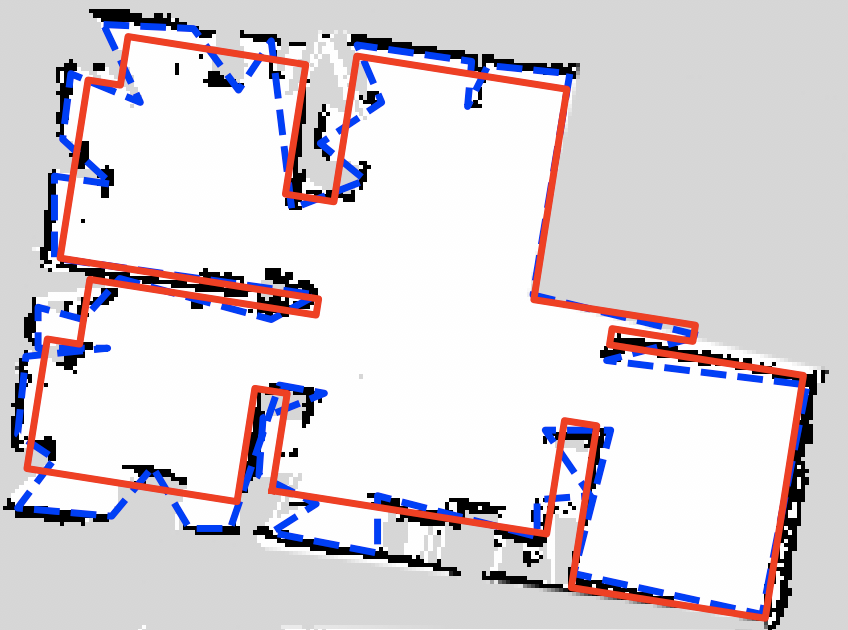}
    \caption{Comparison of Douglas–Peucker (blue) and rectilinear (red) simplifications, overlaid with the greyscale grid obtained the ROS GMapping algorithm.}
    \label{fig:simplification}
\end{figure}

To reduce noise, we simplify the boundary. One standard simplification
method, implemented in OpenCV, is the Douglas-Peucker algorithm
\cite{DouglasPeucker-Simplification}. However, we find that in some
cases this works poorly on the scanned room data. Instead, we use the
simplification algorithm described in \cite{rectilinear} to find the
best fit \textit{rectilinear} polygon with the minimum number of
vertices that remains within a specified tolerance of the original
polygon. More precisely, we find the minimum area bounding rectangle
of the unsimplified polygon to determine the directions of orthogonal
compression, apply \cite{rectilinear}, and discard any
self-intersections to simplify the boundary in linear
time. Fig. \ref{fig:simplification} provides an example of an
occupancy grid and the resulting approximation. Here, we find that
using the rectilinear algorithm produces a very reasonable room
boundary.

\subsection{Waypoint selection}
\label{section:waypoint_selection}
With the floor plan obtained from the mapping algorithm, we now focus on determining the waypoints---that is, we want to choose the specific places within the room that the robot should stop and turn the UVC light on to disinfect the room. 
\subsubsection{Modeling irradiation}
\label{sec:model}

Suppose we turn on the UVC Light at some point $u$ and are interested
in how much energy is irradiated at a point $v$. We measure amount of
irradiation received with the following (standard) assumptions:

\begin{itemize}
    \item $v$ is irradiated by the light only if directly visible from $u$ (we do not consider reflection of light).
    \item The intensity of the light received is proportional to the amount of time that the light is on and inversely proportional to the square of the distance between $u$ and $v$.
    \item If $v$ is at a wall, the intensity is also inversely proportional to the the incidence angle (i.e., vertical angle created  between the normal vector of the wall containing $v$ and vector $st$). If $v$ is not at a wall, just extend the ray from $u$ to $v$ until it hits a wall to obtain the incidence angle.
\end{itemize}

We also assume that the UVC light is a single point whose height is
halfway between the ceiling and the floor. This effectively allows us
to transform the problem into a 2D one: the amount of irradiation
received at the ceiling is the same as the irradiation received at the
floor, and any other horizontal slice will receive more
irradiation. Thus, as long as we sufficiently irradiate the ceiling or floor, we can guarantee full irradiation of the 3D volume.

This assumption is conservative: in practice, the light emanates from a lamp whose shape is similar to traditional fluorescent lamps (a vertical segment would be a more accurate representation). Light emanating from a segment would irradiate more since \textit{(i)} it reduces the distance to $v$ and \textit{(ii)} makes the incidence angle smaller. Thus, using a segment light increases the amount of irradiation received at $v$.\footnote{An ideal placement would locate the light so that it is halfway from the floor and the ceiling. As we will see later, this has an impact on the shadow, but for now we ignore this.}

With the above assumptions, we can model the amount of irradiation received at $v$ from $u$ in a unit of time as:

\begin{equation}
\text{Ir}_u(v) = 
\begin{cases}
\frac{\cos(\alpha_{uv}) P }{d_{uv}^2 + h^2} &\text{if u sees v}\\
0 &\text{otherwise}
\end{cases}
\end{equation}

Here, $\alpha_{uv}$ is the incidence angle between $u$ and $v$ (see
definition above), $P$ is a scaling constant (proportional to the
intensity of the UVC light), $d_{uv}$ denotes the 2D distance between
$u$ and $v$, and $h$ is half of the distance between the floor and
ceiling. We say that a point $v$ is visible from a point $u$ if (1)
the segment connecting the the two points does not intersect the
polygon boundary, and (2) $d_{uv}$ is larger than the shadow distance
(i.e., the UVC light does not see the points directly below it). This
shadow distance is based on some of the characteristics of the robot
and the light source.

\subsubsection{Waypoint selection as an LP problem}
The above formula computes the irradiation from $u$ to $v$ for a single unit of time. In general, if the light is on for $t$ units of time, then $v$ would receive $t\cdot \text{Ir}_u(v)$ irradiation. Note that, for any fixed pair $u$ and $v$, the amount of irradiation is proportional on the time that the robot spends there. 

In order to make sure we fully disinfect the whole floor plan, we need to distinguish between the region we need to guard (the whole floor plan, denoted by $\mathcal{R}$) and the areas within the room that we can reach (denoted by $\mathcal{G}$). $G$ can be obtained by removing from $\mathcal{R}$ all the points that are within a distance $d$ of an obstacle, where $d$ is the radius of the robot.

We apply a grid discretization of both regions by drawing horizontal and vertical lines that are $\varepsilon$ units apart, partitioning $\mathcal{G}$ and $\mathcal{R}$ into grid cells. Let $G^{\mathcal{G}}$ be the set containing the centers of all grid cells that lie within $\mathcal{G}$. We consider points of $G^{\mathcal{G}}$ as potential waypoints for the robot.

We discretize $\mathcal{R}$ in a similar fashion. This gives us a set of points $G^{\mathcal{R}}$ that we need to guarantee that are disinfected. A point is disinfected if it has received enough irradiation (say, received at least $r$ units of irradiation).

This leads to a natural linear problem formulation: for any pair (waypoint, place to disinfect), measure how much is the point irradiated and accumulate over all possible waypoints. Globally, we want to minimize the time spent irradiating while at the same time make sure that all points are guarded. 

Let $t_u$ be the time that we spent at waypoint $u\in G^{\mathcal{G}}$.
The waypoint selection problem is formalized as follows:

\begin{gather*}
\begin{aligned}
&\min \sum_{u \in G^{\mathcal{G}}} t_u 
\text{s.t.  } t_u \geq 0  \quad \forall u \in G^\mathcal{G}\\
&\sum_{u \in G^{\mathcal{G}}} t_u\text{Ir}_{}(v) \geq r \quad \forall v \in G^\mathcal{R}
\end{aligned}
\end{gather*}

Note that because of the robot limitations, this often yields infeasible instances. This is simply because we would often have some corner of $G^{\mathcal{R}}$ that is not seen by any point of $G^{\mathcal{G}}$. Thus, we remove from $G^{\mathcal{R}}$ points that are not seen at all from $G^{\mathcal{G}}$. The percentage of points that need to be removed is also used as a measure of quality of the algorithm (the goal being disinfecting as close to $100\%$ of the room as possible). See Section~\ref{sec:experiments} for more details.

The above formulation guarantees that all points of $G^\mathcal{R}$ (visible from $G^\mathcal{G}$) are disinfected, but even when all points of $G^{\mathcal{R}}$ are visible, this does not guarantee that the whole room is disinfected. Indeed, when two consecutive grid points are seen by different waypoints we cannot guarantee that intermediate points are seen by either waypoint.

In the following, we provide two different modifications to the algorithm that would fix this issue.

\subsubsection{Guaranteed disinfection}
Our first approach modifies the definition of irradiance. We say that the ``pessimistic irradiance'' between a waypoint $u$ and a room point $v$ is the \textit{minimum} irradiance between $u$ and any point in the grid cell containing $v$. A feasible solution to the pessimistic irradiance LP problem will guarantee to disinfect, not only the discrete grid $G^{\mathcal{R}}$, but also the neighborhoods of all such points. By the construction of the grid cells, the union of all such neighborhood will be $\mathcal{R}$.

Alternatively, we can scale up the waiting times to guarantee that every point in the room receives the minimum level of irradiance. We locate the globally dimmest point in the room with respect to the waiting times and then compute a scale factor to ensure that point receives the minimum irradiance. Balakrishnan et al. \cite{branchbound} give a 
\textit{branch-and-bound} algorithm for finding global minima using upper and lower bounds on the function's minimum within a subregion of the search space.
We find an upper bound on minimum irradiance in a region of a room by choosing an arbitrary point within the region and computing its irradiance. For a lower bound, we compute the pairwise irradiance between the guard points and vertices of the region, choose the minimum irradiance over the vertices for each guard point, and sum over the guard points. We use a \textit{triangulate-and-bound} variant of Balakrishnan's algorithm: we triangulate the room and split the triangles along their longest edge until, for each triangle, all vertices are covered by the same guard point (accounting for shadows if desired). For the branching procedure, we continue to split triangles in this manner. At the cost of a small decrease in coverage, we can achieve tighter bounds by allowing the removal of triangles with small areas or loose lower bounds from the search space.

Either of these two modifications guarantees that the visible region of the volume is disinfected. Walls of the region are also disinfected by the original LP formulation: while we only focus on disinfecting the \textit{lowest} points along the wall, this is sufficient to disinfect the whole wall. Indeed, as distance and incidence angle are both maximized when the point is at the bottom of the wall, irradiance is minimized by points on the bottom of the wall. Thus, any solution that disinfects the bottom of the wall must also disinfect the entire wall.

It can be shown that as the discretization parameter $\varepsilon$ approaches 0, the ``pessimistic irradiance'' algorithm converges to the minimum time needed to disinfect the room. The ``branch-and-bound'' algorithm converges in certain rooms, and we conjecture that it will perform reasonably in the general case.

\subsection{Route planning}
Once the waypoints (and stopping times for each of them) have been determined, we need to find an efficient way to visit all of them. This is a variation of the traveling salesman problem where weights are geodesics (i.e., the weight between any two points is the shortest paths within a simple polygon). We compute the geodesic distance between the waypoints by running Dijkstra, and then compute an approximation of the geodesic TSP that are available to the robot. In practice, the transport times are so small that they are overshadowed by the time spent at the waypoints.

In addition, we have to consider the orientation of the robot (i.e., it can only move forward in one specific direction, and whenever we need to change direction it must rotate). Good approximation solutions are known for these kind of problems, but unfortunately they run in exponential time~\cite{curvature}. The simpler solution used is to ensure waypoints do not place the robot too close to any obstacles (which is already being calculated) and to then allow the robot's path planner determine the best way to get in and out of that position. 
\section{Evaluation Methods}
\label{section:evaluation_methods}
We evaluate a solution by looking at \textit{(i)} the percentage of the room that is disinfected\footnote{Some corners could not be disinfected in our instances} while at the same time \textit{(ii)} minimizing the time required to complete the disinfection. To establish a baseline for comparison, we propose a naive algorithm simulating a stationary disinfection routine: choose a single point near the center of $\mathcal{G}$\footnote{Specifically, the midpoint of the largest segment of the horizontal bisector of the polygon. This is implemented in the \texttt{representative\_point} function of the Shapely library} and wait as long as necessary to completely disinfect the farthest point of the room that it can see. Since rooms are rarely starshaped, it will be unlikely that a single point will be able to guard the room. We expect this naive algorithm to perform terribly (in both of our optimization criteria); we mainly use it as a baseline for comparison purposes.

Note that our disinfection coefficient is fairly high compared to similar experiments. Note that in our experiments, a point must be irradiated with $120600 \mu Ws/cm^2$, which is a fairly high threshold (say, when compared to the $12000-22000 \mu Ws/cm^2$ threshold required in a similar experiment~\cite{trud_eval}.). We do this because SARS-CoV-2 has been shown to be fairly resilient to irradiation~\cite{Derraik}. In any case, we note that the value of the parameter itself has little impact on the results: coverage is unaffected, and a change in the coefficient would have a linear impact in time needed by both our baseline (stationary) and our algorithm. 

We can also find an upper bound on the disinfection percent and a lower bound on the time required to achieve that percent for \textit{any} path the robot can take. We use a similar idea to the ``pessimistic irradiance'' algorithm, applied in reverse: we consider the \textit{best} possible irradiance possible between a point in the grid cell around the robot location and a point in the room. Define the ``optimistic irradiance'' between a point $u \in G^\mathcal{G}$ and a point $v \in G^\mathcal{R}$ to be the \textit{maximum} irradiance between any point in the grid cell containing $u$ and the point $v$. As the grid cells cover $\mathcal{G}$ (ie. all possible locations the robot can be), running the linear program with ``optimistic irradiance'' constraints will produce an upper bound on the disinfection percentage and a lower bound on the disinfection time. As the discretization parameter $\varepsilon$ approaches $0$, these bounds tighten.

\section{Experiments}\label{sec:experiments}

\begin{table}[tb]
\caption{Comparison of different strategies}
\label{table:case_study_shadow}
\begin{center}
\begin{tabular}{|c|c|c|c|c|c|c|}
\hline
\multicolumn{7}{|c|}{Algorithms that consider robot shadow} \\ \hline
&\multicolumn{2}{|c|}{\textbf{Branch \& Bound}}
&\multicolumn{2}{|c|}{\textbf{Pessimistic}}
&\multicolumn{2}{|c|}{\textbf{Lower Bound}}\\
\cline{2-7}
\textbf{$\varepsilon$} & \textbf{\textit{Time}} & \textbf{\textit{Disinf.}} & 
\textbf{\textit{Time}} &
\textbf{\textit{Disinf.}} &
\textbf{\textit{Time}}&
\textbf{\textit{Disinf.}}\\
& \textbf{\textit{(s)}}& \textbf{\textit{\%}} &
\textbf{\textit{(s)}}& \textbf{\textit{\%}} &
\textbf{\textit{(s)}}& \textbf{ \textit{\%}} \\\hline
1.0& 112389 & 99.29 & 74804 & 91.54 & 17444 & 99.95\\
0.5& 107893 & 95.10 & 58076 & 98.22 & 29434 & 99.90\\
0.3& 88573 & 99.61 & 49211 & 99.14 & 29034 & 99.87\\
0.2& 133928 & 99.60 & 39393 & 99.48 & 29935 & 99.85\\
\hline \hline
\multicolumn{7}{|c|}{Algorithms that ignore robot shadow} \\ \hline
&\multicolumn{2}{|c|}{\textbf{Branch \& Bound}}
&\multicolumn{2}{|c|}{\textbf{Pessimistic}}
&\multicolumn{2}{|c|}{\textbf{Lower Bound}}\\
\cline{2-7}
\textbf{$\varepsilon$} & \textbf{\textit{Time}} & \textbf{\textit{Disinf.}} & 
\textbf{\textit{Time}} &
\textbf{\textit{Disinf.}} &
\textbf{\textit{Time}}&
\textbf{\textit{Disinf.}}\\
& \textbf{\textit{(s)}}& \textbf{\textit{\%}} &
\textbf{\textit{(s)}}& \textbf{\textit{\%}} &
\textbf{\textit{(s)}}& \textbf{\textit{\%}} \\\hline
1.0& 110969 & 99.29 & 57479 & 93.00 & 17444 & 99.95\\
0.5& 107556 & 95.10 & 52921 & 98.47 & 29421 & 99.90\\
0.3& 79854 & 99.61 & 47306 & 99.23 & 28812 & 99.87\\
0.2& 132187 & 99.60 & 37734 & 99.48 & 29602 & 99.85\\
\hline
\end{tabular}
\label{tab1Combined}
\end{center}
\end{table}

\begin{figure}[tb]
    \centering
    \includegraphics[width=.7\textwidth]{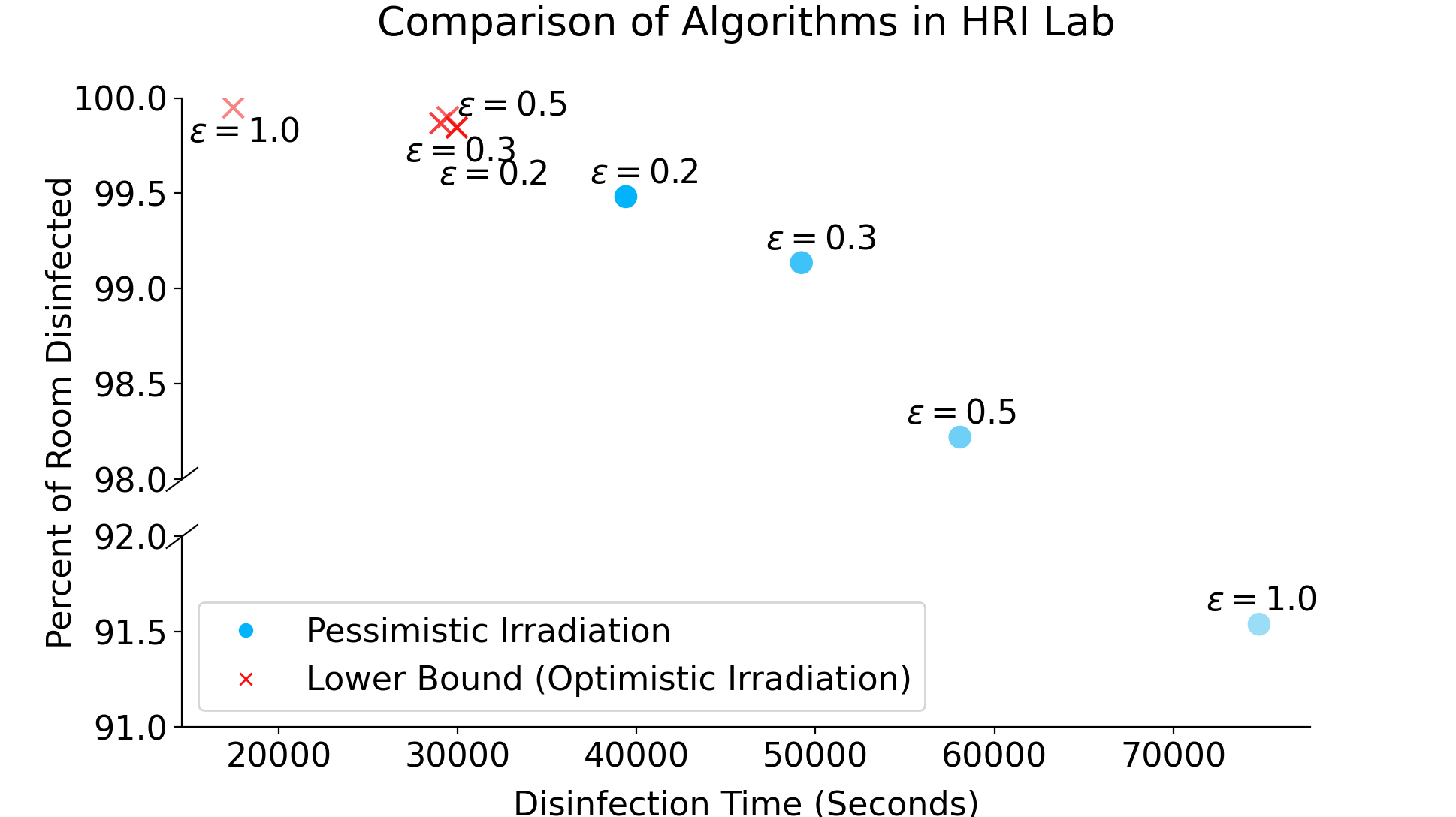}
    \caption{Bicriteria evaluation of the different algorithms on the HRI Lab instance.}
    \label{fig:comparison_in_hri}
\end{figure}

\begin{table*}[tb]
\caption{Summary of Performance In All Rooms, $\varepsilon = 0.2$}
\begin{center}
\begin{adjustbox}{width=0.86\textwidth}
\begin{tabular}{|c|c|c|c|c|c|c|c|c|}
\hline
&\multicolumn{2}{|c|}{\textbf{Naive}}
&\multicolumn{2}{|c|}{\textbf{Branch \& Bound}}
&\multicolumn{2}{|c|}{\textbf{Pessimistic Irradiance}}
&\multicolumn{2}{|c|}{\textbf{Lower Bound}}\\
\cline{2-9}
\textbf{Room Name} & & \textbf{\textit{Percent}} & &
\textbf{\textit{Percent}} & &
\textbf{\textit{Percent}} & & \textbf{\textit{Percent}}\\
& \textbf{\textit{Time (s)}}& \textbf{\textit{Disinfected}} &
\textbf{\textit{Time (s)}}& \textbf{\textit{Disinfected}} &
\textbf{\textit{Time (s)}}& \textbf{\textit{Disinfected}} & \textbf{\textit{Time (s)}}& \textbf{\textit{Disinfected}} \\
\hline
HRI Lab  & 99937 & 41.65  & 133928 & 99.60 & 39393 & 99.48 & 29935 & 99.85\\
Room 2510  & 41403 & 85.15 & 200016 & 97.28 & 13546 & 98.60 & 12902 & 99.17\\
Room 2530  & 40845 & 94.19 & 35237 & 99.86 & 13958 & 99.22 & 16684 & 100.00\\
Room 2540\footnotemark  & 32228 & 58.07 & 79637 & 71.02 & 73654 & 72.20 & 67905 & 74.50\\
Room 2560 & 19066 & 57.88 & 31520 & 100.00 & 11854 & 100.00 & 7459 & 100.00\\
Room 2910  & 42857 & 84.74 & 12634 & 97.72 & 16323 & 98.18 & 9849 & 98.87\\
\hline
\end{tabular}
\end{adjustbox}
\label{table:summary_all_rooms}
\end{center}
\end{table*}

\footnotetext{The ``pessimistic irradiance'' approach is unusually bad in this room due to a failure of the room boundary simplification algorithm, which in this case slightly exaggerates an obstacle and makes half the room inaccessable to the robot.}

Six rooms were mapped and served as test data. We closely
examine one room, the ``HRI lab'', as a case study for the various
parameters of the proposed algorithm.
\footnote{Code \& data available at \href{https://github.com/pjrule/covid-path-planning}{github.com/pjrule/covid-path-planning}.}

To establish a baseline we compute the amount of time that a stationary light would need when placed in the position that would maximize the coverage (i.e., disinfect the largest possible portion of the room). Just to give an example, a naive stationary solution (described in~\ref{section:evaluation_methods}) yields a baseline disinfection time
of $99,937$ seconds to disinfect $41.65\%$ of the room. A low disinfection percent is caused by the nonconvexity of the lab (i.e., walls and doors create many visibility constraints). Table
\ref{table:case_study_shadow} and Fig. \ref{fig:comparison_in_hri}
give a comparison between the two algorithms described in
\ref{section:waypoint_selection} and the lower bound described in
\ref{section:evaluation_methods}, run with various values of the
$\varepsilon$ discretization parameter. Notice that as $\varepsilon$
decreases, the solution times and disinfection percents achieved by
the algorithms appear to quickly converge to the lower bound, as expected. The best
solution time achieved is 2.5 times faster than the naive solution and
disinfects twice as large an area. The branch-and-bound algorithm is
competitive for some values of $\varepsilon$, though it is highly
sensitive to parameterization and therefore does not always produce
tight bounds. The data suggest the ``pessimistic irradiance'' algorithm often converges quickly to optimal; branch-and-bound appears to converge more slowly or not at all.

The effect of the robot's shadow was also investigated. Due to the
physical interference of the robot, the UVC light cannot disinfect
the area directly below it. As this is the point where distance is
minimized, one might worry that this negatively affects solution
times. Table \ref{tab1Combined} shows the results of the simulations
run for variations that consider or ignore the visibility. At
$\varepsilon = 0.2$, the solution accounting for robot shadow only
takes $4\%$ longer than the solution that ignores shadow while
disinfecting the same area: the effect is negligible.

Table \ref{table:summary_all_rooms} summarizes the performance of the various algorithms across all rooms at the discretization parameter $\varepsilon = 0.2$. Similar to the case study, the ``pessimistic irradiance'' algorithm performs well relative to the lower bound and is a significant improvement over the naive approach.

\subsection{Example run in the office environment}

An example run of the algorithm fully integrated with a robotic platform
is demonstrated~\footnote{\label{footnote:vidlink}Video of the demonstration can be found at \href{https://hrilab.tufts.edu/movies/autonomous_disinfection.mp4}{https://hrilab.tufts.edu/movies/autonomous\_disinfection.mp4}. Exposure times are reduced to 1 second for succinctness.}.
First, the operator designates a zone to be disinfected by mapping
it through teleoperation. When the
operator is satisfied, the map can be reused for
disinfection and path planning (assuming the
environment does not drastically change).

Environment data inis provided to an implementation of the algorithm which produces the waypoints and exports them as location/time pairings in a CSV file. These pairings are read on startup of the robot software stack, and are referenced when the ``begin disinfection'' task is called. This task was triggered manually when testing, but scheduled approaches are also viable (assuming safety steps are taken to ensure no one is harmed by direct UVC exposure).

When the environment is ready for disinfection, the disinfection task is triggered and the robot moves to each point, ensuring that it remains at each point no less than the time specified by the algorithm. The light remains on between waypoints, because as outlined by \ref{section:method} the error introduced by doing so is small and the consequences of additional exposure are more positive than negative.

The robot is placed just outside an empty pre-mapped office environment for which waypoints have been computed using the algorithm described, and then navigates into the room, turns on the lamp, and completes the loop specified by the waypoints. Some points were mapped as a result of being able to see them through glass doors, and so could not be navigated to: these points were removed for demonstration after having seen them handled sensibly (displaying an error but continuing to disinfect remaining achievable regions). To prevent accidental human UV-C exposure, the doors to the office are locked and labeled. The operator is in another room behind a closed door, which is where the robot tether leads for both power and monitoring purposes. All robot hardware and navigation control is handled on the platform.

\section{Conclusion and Future Work}

We have developed a low-cost autonomous robot platform together with a
planning algorithm for UVC irradiation tasks and demonstrate effectiveness in
simulation and hardware.
While the current approach focused on using a floor plan for
disinfection, future work will address the more difficult problem of
computing the minimum necessary exposure time from 3D data, perhaps as
a continuous path, which would allow the disinfection platform to use
its travel time as disinfection time
Additionally, improvements can be made to the platform, as
the safety restrictions imposed as a result of COVID-19 impeded
manufacturing and in turn forced design choices which could otherwise be
avoided.

\section{Acknowledgements}
This project was funded in part by a Tufts COVID-19 Rapid Response Award. The authors would like to thank Diane Souvaine and the other members of the Computational Geometry laboratory at Tufts University for their suggestions and their helpful discussions. 
 
\small
\bibliographystyle{abbrv}
\bibliography{cfree}

\end{document}